\documentclass[lettersize,journal]{IEEEtran}

\usepackage[utf8]{inputenc}
\usepackage[T1]{fontenc}
\usepackage{amsmath,amsfonts,amssymb}
\usepackage{algorithmic}
\usepackage{algorithm}
\usepackage{array}
\usepackage[caption=false,font=normalsize,labelfont=sf,textfont=sf]{subfig}
\usepackage{textcomp}
\usepackage{stfloats}
\usepackage{url}
\usepackage{verbatim}
\usepackage{graphicx}
\usepackage{grffile}
\usepackage{cite}
\usepackage[dvipsnames]{xcolor}
\usepackage{booktabs}
\usepackage{multirow}
\usepackage{tabularx}
\usepackage{makecell}
\usepackage{threeparttable}
\usepackage{float}
\usepackage{caption}
\usepackage{wrapfig}
\usepackage{placeins}
\usepackage{paralist}
\usepackage{pifont}
\usepackage{microtype}
\usepackage{nicefrac}
\usepackage{booktabs}
\usepackage{multirow}
\usepackage{array}
\usepackage{makecell}
\usepackage{makecell}
\graphicspath{{figures/}}
\newcommand{\NOTE}[1]{\textcolor{red}{}}

\newcommand{\CUT}[1]{}

\definecolor{MyGreen}{RGB}{0,128,0}
\definecolor{MyRed}{RGB}{220,20,60}

\newcolumntype{Y}{>{\raggedright\arraybackslash}X}
\newcolumntype{C}{>{\centering\arraybackslash}X}

\makeatletter
\let\@oldmaketitle\@maketitle% Store \@maketitle
\renewcommand{\@maketitle}{\@oldmaketitle % Update \@maketitle to insert...
  \centering
  \includegraphics[width=0.9\textwidth]{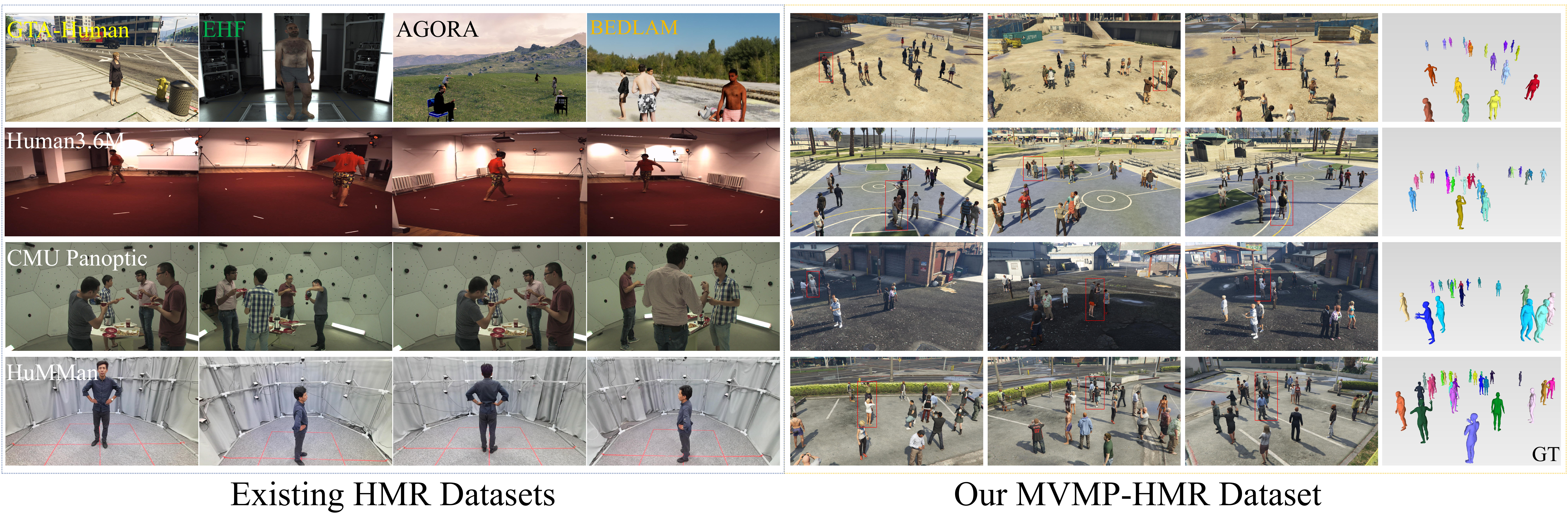}
 \setcounter{figure}{0} 
 \captionof{figure}{Comparison between representative existing single-view and multi-view HMR datasets with our proposed MVMP-HMR dataset.
The first row on the left shows single-view HMR datasets: GTA-Human~\cite{GTA-Human}, EHF~\cite{EHF}, AGORA~\cite{Agora}, and BEDLAM~\cite{Bedlam}.
The following three rows present multiview datasets, including Human3.6M~\cite{Human3.6m}, CMU Panoptic~\cite{joo2015panoptic}, and HuMMan~\cite{cai2022humman}.
In contrast, MVMP-HMR provides synchronized multiview images and corresponding ground-truth SMPL-X meshes captured in large outdoor environments with substantially more subjects.
Red boxes highlight regions with severe inter-person occlusions.} 
 \label{fig:dataset}
 \vspace{-0.8cm}
 }
\makeatother

\begin{document}

\title{Multiview Multi-Person Human Mesh Recovery Under Large Scenes with Occlusions}

\author{Qi~Zhang,~\IEEEmembership{Member,~IEEE,}
        Tao~Yu, %~\IEEEmembership{Senior Member,~IEEE}
        Jiechao~He, %~\IEEEmembership{Senior Member,~IEEE}
        Antoni B. Chan,~\IEEEmembership{Senior Member,~IEEE}\\
        and~Hui~Huang*,~\IEEEmembership{Senior Member,~IEEE}
\thanks{Qi Zhang, Tao Yu, Jiechao He, and Hui Huang are with the Guangdong Provincial Key Laboratory of Visual Media and Multidimensional Intelligence, College of Computer Science and Software Engineering, Shenzhen University, China. Antoni B. Chan is with City University of Hong Kong, China.
E-mail: qi.zhang.opt@gmail.com, 2310275022@email.szu.edu.cn, 2510104004@email.szu.edu.cn, abchan@cityu.edu.hk, hhzhiyan@gmail.com} 
% <-this % stops a space
\thanks{*Corresponding author.}
\thanks{Manuscript received xxx; revised August xxx.}
}

\markboth{~Vol.~xx, No.~xx, 2026}%
{First Author \MakeLowercase{\textit{et al.}}: Multiview Multi-Person Human Mesh Recovery Under Large Scenes with Occlusions}

\maketitle

\begin{abstract}
Human mesh recovery (HMR) aims to recover 3D human meshes from images. 
Most existing HMR benchmarks and methods focus on either multi-person reconstruction from a single view or single-person reconstruction from multiple views, where the number of subjects and the scene scale are relatively limited.
Such settings are insufficient for real-world applications with large scenes and severe inter-person occlusions. To address this limitation, we introduce a large-scale synthetic benchmark for multiview multi-person HMR, termed MVMP-HMR.
The proposed dataset contains 15 complex scenes with up to 50 camera views and 30 interacting persons, featuring large spatial coverage and severe occlusions, which significantly increases the difficulty of human mesh recovery. Based on this benchmark, we further propose a multiview multi-person whole-body human mesh recovery model, referred to as MVMP-HMR model. The model first fuses multiview features into a scene-level 3D feature volume, and then leverages pelvis joints predicted by a 3D pose estimation network to extract person-specific queries from the 3D feature volume. These human queries are cross-attended with the 3D feature volume and integrated to decode each person's 3D mesh. Moreover, we introduce two novel losses--the orientation loss and the 3D joint density loss--to alleviate orientation and pose ambiguities under severe occlusions. Experiments demonstrate that existing state-of-the-art HMR methods struggle on the proposed MVMP-HMR benchmark, while our method consistently outperforms prior SOTAs in large-scale scenes with severe occlusions.
\end{abstract}

\begin{IEEEkeywords}
Multiview, Multi-person, HMR, Large scenes.
\end{IEEEkeywords}

% \begin{figure}[!t]
%   \centering
%   \includegraphics[width=1.0\linewidth]{figures/dataset.pdf}
%   \caption{Comparison between representative single-view and multi-view HMR datasets with our MVMP-HMR dataset.
% The first row on the left shows single-view HMR datasets (GTA-Human~\cite{GTA-Human}, EHF~\cite{EHF}, AGORA~\cite{Agora}, and BEDLAM~\cite{Bedlam} from left to right).
% The following three rows present multiview datasets, including Human3.6M~\cite{Human3.6m}, CMU Panoptic~\cite{joo2015panoptic}, and HuMMan~\cite{cai2022humman}.
% In contrast, MVMP-HMR (right) provides synchronized multiview images and corresponding ground-truth SMPL-X meshes captured in large-scale outdoor environments with substantially more subjects.
% Red boxes highlight regions with severe inter-person occlusions.}
%   \label{fig:dataset}
% \end{figure}

%\zq{(Draw a figure comparing the single-HMR and our dataset.)}
% \begin{figure}[h]
%   \centering
%   \vspace{-0.4cm}
%   \includegraphics[width=0.8\linewidth]{figures/dataset.pdf}
%    \vspace{-0.2cm}
%   \caption{Comparison between single-view HMR datasets and our proposed MVMP-HMR dataset. The left shows images from GTA-Human \cite{GTA-Human}, EHF \cite{EHF}, AGORA \cite{Agora}, and BEDLAM \cite{Bedlam} datasets from top to bottom. Our multiview images and ground truth meshes are shown on the right, containing larger scenes and more persons. The red box indicates areas with severe occlusions.}
%   \label{fig:dataset}
%   \vspace{-0.7cm}
% \end{figure}

\section{Introduction}
\label{sec:intro}

Human mesh recovery (HMR) predicts the human 3D meshes from images or image crops, which has important applications in autonomous driving, digital games, or AR/VR, \textit{etc}. Most existing HMR methods focus on recovering human meshes for scenes with a quite limited people number (usually $<15$ in total), either with a single person from single images or multi-crops, or multi-persons from single images.  Besides, the evaluation benchmarks used in the latest methods are usually under small scenes, with few occlusions (see Fig. \ref{fig:dataset} left). This is not practical for real-world applications where there might be massive crowds in large scenes with severe occlusions. Thus, the existing HMR methods have not been evaluated under more complicated conditions with both larger human numbers and more severe occlusions, whose performance is not ensured.

\begin{figure*}[t]
  \centering
  \includegraphics[width=0.85\linewidth]{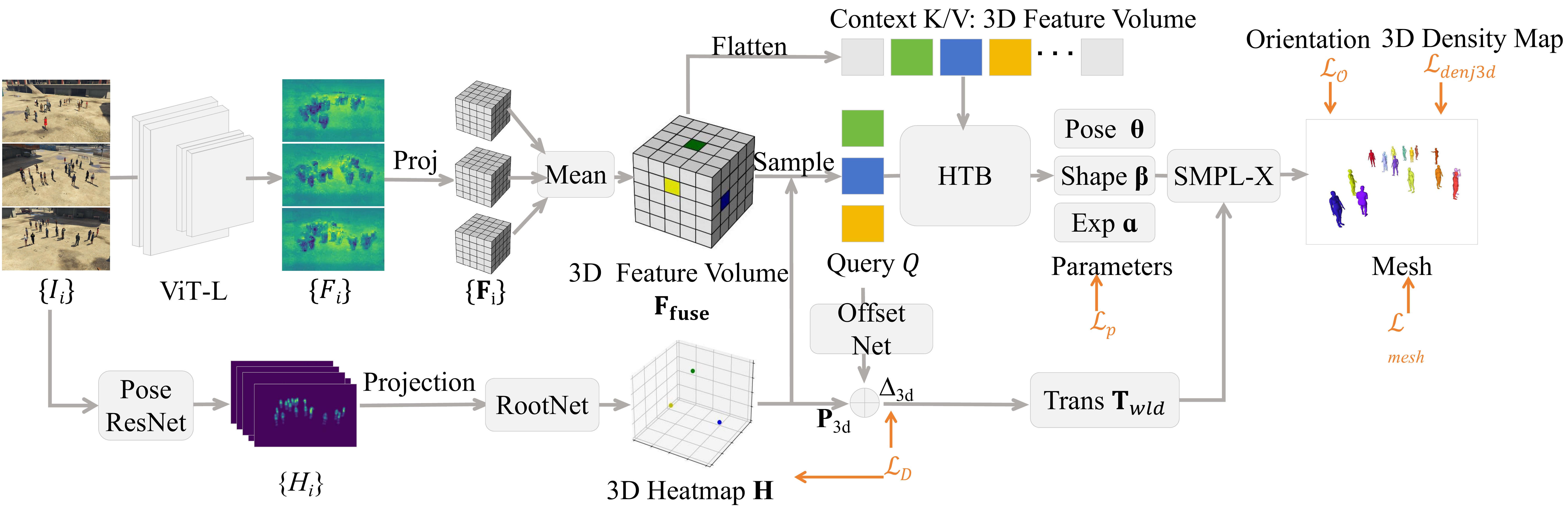}
  \vspace{-0.25cm}
  \caption{Overview of the proposed MVMP-HMR pipeline.
    Single-view features are first extracted using a ViT backbone and projected into a shared 3D space to form a scene-level feature volume. Person-specific queries are then derived from a 3D pose estimation branch and fed into a human transformer block (HTB) to decode SMPL-X parameters. We further introduce an orientation loss $\mathcal{L}_{\mathcal{O}}$ and a 3D joint density loss $\mathcal{L}_{\mathrm{denj3d}}$ to improve orientation and pose accuracy.}
  % \caption{The pipeline of our proposed MVMP-HMR method. We first extract single-view features with a ViT backbone, and then the single-view features are projected to the 3D space and averaged to obtain the 3D feature volume of the whole scene. Finally, with the joints outputted from a 3D pose estimation branch (at the bottom), we extract human queries for each person and feed them into a human transformer block (HTB) for 3D decoding and SMPL-X parameters prediction. In addition to losses previously used in single-view HMR SOTAs, we also put forward two novel losses, \textit{eg.}, orientation loss $\mathcal{L}_\mathcal{O}$ and 3D joint density loss $\mathcal{L}_{denj3d}$, for better orientation and pose accuracy in meshes.}
  \label{fig:pipeline}
  \vspace{-0.3cm}
\end{figure*}

To solve the problem and extend the HMR task to more complicated scenes, in this paper, we present MVMP-HMR (as in Fig. \ref{fig:pipeline}), a novel model for multi-person whole-body human mesh recovery from multiview images, which fuses multiview clues to handle the severe occlusions in large scenes with more humans.
%Specifically, Multiview-HMR fuses multiple single-view features via a novel depth-guided projection and fusion technique, then predicts the SMPL parameters for obtaining multiple human meshes in 3D space.
Specifically, MVMP-HMR extracts single-view features and projects them to the 3D space, and then the projected multiview features are averaged to obtain a complete 3D feature volume for the whole scene. Besides, a 3D pose estimation branch is adopted to predict the pelvis joint location of each person, and the predicted pelvis joint is used to acquire the human queries by sampling at the locations from the previously fused 3D feature volume. Then the human queries and the 3D feature volume are both fed into the human transformer block (HTB) where both are fused via cross-attention layers. Finally, the output of HTB is decoded to regress the SMPL-X parameters.
%\yt{Deformable attention layer is used aggregate the feature volume after the feature projection}

To deal with the human orientation and pose ambiguities in the predicted SMPL-X parameters under the occluded scenes, in addition to common parameter regression losses used in Multi-HMR \cite{baradel2024multi}, we put forward two novel losses: the \textbf{orientation loss} and the \textbf{3D joint density loss}. The orientation loss $\mathcal{L}_\mathcal{O}$ is the supervision of the human mesh's orientation in the real-world coordinates. The 3D joint density loss $\mathcal{L}_{denj3d}$ supervises the 3D joints in the predicted human mesh via 3D joint density maps instead of direct joint coordinate regression. Both provide stronger supervision in the 3D space and handle the orientation and pose ambiguities in the MVMP-HMR task better, further enhancing the model performance (see results in Sec. \ref{section:ablation}).
Furthermore, we introduce a large-scale synthetic multiview multi-person HMR dataset with more subjects, camera views, and scene variations than existing datasets.

{In summary, the contributions of the paper are:}
\begin{compactitem}
    \item We propose a large synthetic MVMP-HMR dataset for studying the multiview multi-person HMR task. Compared to existing datasets, the proposed one contains larger scenes with severe occlusions, which are more suitable for real-world applications.
    %As far as we know, this is the first study on the multiview multi-person HMR task under large scenes with severe occlusions. No existing research has focused on the issue in the HMR area. Besides, we propose a large MVMP-HMR dataset for studying the topic. 
    \item  We propose the MVMP-HMR model, which is an end-to-end multiview multi-person HMR model for reconstructing multiple persons with multiple views under large scenes. In addition, we introduce two novel losses to guide 3D human orientation and body pose estimation, leading to improved performance on our dataset.
    \item  Experimental results show that existing methods struggle on the  multiview multi-person HMR benchmark as well as other real-world datasets. In contrast, our method consistently outperforms state-of-the-art HMR methods and multiview 3D human pose estimation approaches.
\end{compactitem}

\section{Related Work}
\label{sec:relatedWork}

\textbf{Single-person HMR.}
Human mesh recovery (HMR) predicts the human 3D meshes from images. 
The early HMR methods were based on optimization, and they were easily stuck at local minima \cite{hasler2010multilinear, lin2023one, moon2022accurate, EHF, liu2024deep}. %[14,31,42,47]
Instead of estimating the human meshes as in 3D reconstruction, \cite{kanazawa2018end} proposed to predict SMPL parameters of the shape and 3D joint angles to represent human meshes from a cropped image. 
SMPLify-X \cite{EHF} followed SMPLify to estimate the 2D joints and optimize model parameters to fit them, and then improved over SMPLify with a new DNN trained on a larger dataset.
In addition, many regression-based methods were proposed \cite{cai2023smpler, choutas2020monocular, feng2021collaborative, moon2022accurate, rong2021frankmocap,zhang2023pymaf, zhou2021monocular, su2025sat, wang2025prompthmr}, 
which is focused on single-person estimation. 
%PyMAF-X \cite{zhang2023pymaf} presented a novel framework for accurate full-body mesh regression from monocular images, using a Pyramidal Mesh Alignment Feedback (PyMAF) loop to enhance alignment precision.
Furthermore, many methods tried to utilize multi-crops to enhance the HMR performance
\cite{choutas2020monocular, feng2021collaborative, moon2022accurate, lin2023one, cai2023smpler, yu2022multiview, huang2021dynamic}.
To alleviate depth ambiguity and improve geometric consistency, multi-view single-person HMR methods~\cite{matsubara2025heatformer, li2024human} exploit synchronized observations from multiple cameras. HeatFormer \cite{matsubara2025heatformer} performs joint multi-view neural optimization, refining SMPL parameters by enforcing cross-view agreement through heatmap-based supervision. In summary, single-person HMR is limited to images with few persons, making it impractical for real-world scenarios with multiple people, larger scenes, and occlusions.
% In addition, HeatFormer \cite{matsubara2025heatformer} is a neural optimization method based on 2d heatmap generated from SMPL parameters. 
% \textit{In summary, single-person HMR is limited to images with few persons, making it impractical for real-world scenarios with multiple people, larger scenes, and severe occlusion.

\textbf{Multi-person HMR.}
Compared to single-person HMR, multi-person HMR \cite{choi2022learning, goel2023humans, kolotouros2019learning, qiu2022dynamic, zhang2021pymaf}
need to predict the human meshes of multiple persons in the images. Multi-person HMR usually adopts a two-stage procedure: detect all humans in the image first \cite{he2017mask, liu2016ssd, redmon2016you},  
and then perform HMR \cite{kim2023sampling, ma2023d, yoshiyasu2023deformable, zheng2023potter} for each detected person with crops. The two-stage process is not end-to-end and the occlusion in images may hurt the human detection accuracy, thus limiting the whole pipeline's performance.
In contrast, single-stage methods have also been proposed \cite{sun2021monocular, qiu2023psvt, sun2022putting}.
Recent methods Multi-HMR \cite{baradel2024multi} and AiOS \cite{sun2024aios} adopted the DETR architecture for multi-person human mesh recovery.
Multi-HMR \cite{baradel2024multi} detects 2D people locations using features of a ViT backbone and predicts their whole-body pose, shape, and 3D location using a cross-attention module.
AiOS \cite{sun2024aios} performs human localization and SMPL-X estimation in a progressive manner, which consists of body localization, body refinement, and a whole-body refinement stage to regress SMPL-X parameters. Beyond mesh recovery, related multi-view human analysis includes sparse-keypoint pose estimation \cite{zhang20204d,dong2021fast,zhang2022voxeltrack,liao2024multiple,wang2021mvp}, avatar reconstruction \cite{lu2024avatarpose,lee2025geoavatar}, and recent multi-person motion capture \cite{jia2026ram,cuevas2026mamma}. Even though existing multi-person HMR methods can accurately estimate human meshes for several persons in single images, they are only evaluated on small scenes containing a small number of persons, eg, $<15$. It is not clear whether they can be applied to scenes with larger sizes and severe occlusions. Thus, we propose the MVMP-HMR dataset and model, which fuses multiple camera views to deal with severe occlusions in crowded scenes.
% \textit{As far as we know, this is the first study for multi-person HMR with multiviews, and we also propose a large synthetic MVMP-HMR dataset, which shall advance the HMR task to more complicated conditions.}

\textbf{HMR and 3D HPE Datasets.}
Numerous datasets have been proposed for Human Mesh Recovery (HMR) and related 3D human tasks (e.g., 3D Human Pose Estimation, HPE). 
\textit{Single-view HMR datasets} such as GTA-Human~\cite{GTA-Human}, AGORA~\cite{Agora}, BEDLAM~\cite{Bedlam}, and EHF~\cite{EHF} provide SMPL-family annotations but rely on monocular capture. As a result, they predict meshes in camera coordinates and inherently suffer from depth ambiguity. Moreover, most scenes contain only one or a small number of subjects and are generated or captured in relatively simple environments, limiting their applicability to large-scale crowded scenarios.
\textit{Multiview HMR datasets} leverage synchronized camera arrays to improve geometric consistency. Datasets such as  HuMMan~\cite{cai2022humman}, Mpi\_inf\_3dhp~\cite{mehta2017monocular}, and Hi4D~\cite{yin2023hi4d} provide multiview supervision with SMPL or SMPL-X fitting. Nevertheless, they primarily focus on single-person reconstruction or small-group interactions in constrained indoor environments, typically involving fewer than five subjects and limited occlusion diversity.
\textit{3D HPE datasets} including Human3.6M~\cite{Human3.6m}, 3DPW~\cite{3DPW}, and CMU Panoptic~\cite{joo2015panoptic} offer accurate 3D joint annotations captured in controlled settings. However, they are generally limited in scene scale (often $\leq 50\,m^2$), environmental variation, and crowd density, resulting in relatively mild inter-person occlusions.
In contrast, our MVMP-HMR dataset is explicitly designed for large-scale multiview multi-person mesh recovery in complex outdoor scenes, featuring up to 30 subjects, expansive areas, dynamic lighting variations, and severe inter-person occlusions. These characteristics make it a more challenging and realistic benchmark for studying the multiview multi-person HMR task.

\begin{figure}[t]
  \centering
  \includegraphics[width=0.98\linewidth]{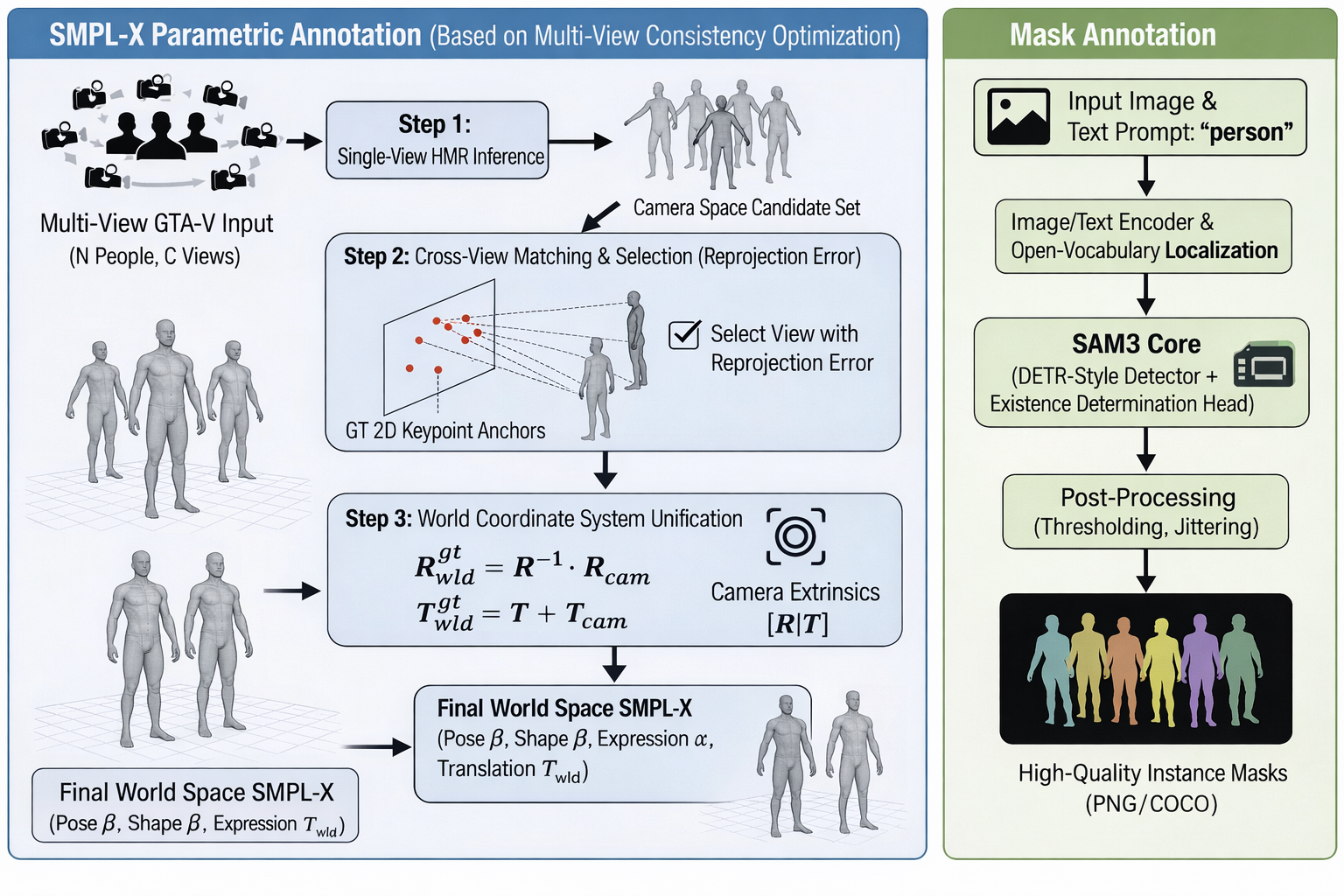}
   \vspace{-0.3cm}
  \caption{Annotation pipeline for MVMP-HMR dataset. %Left: SMPL-X parameters are obtained via multi-view consistency and unified in world coordinates. Right: SAM3-based open-vocabulary segmentation produces high-quality multi-person instance masks.
  }
  \vspace{-0.3cm}
  \label{fig:Annotation}
\end{figure}

\section{MVMP-HMR Dataset}
\label{sec:dataset}
%The proposed MVMP-HMR dataset is a large-scale synthetic dataset designed for multiview multi-person human mesh recovery in complex and crowded scenes.
% We conduct experiments on three datasets: our synthetic MVMP-HMR, Panoptic Studio~\cite{joo2015panoptic}, and Human3.6M~\cite{Human3.6m}.
% MVMP-HMR is collected using the virtual game platform GTA-V and is designed for multiview multi-person human mesh recovery under large scenes with severe occlusions.

\textbf{Dataset Generation.}
MVMP-HMR targets challenging ``in-the-wild-like'' scenes with high crowd density and severe occlusions.
We construct diverse everyday scenes (e.g., basketball courts, factories, and streets), each containing 10--30 people and up to 50 camera views.
Using GTA-V APIs \cite{Wang2019Learning}, we extract (1) RGB frames, (2) 98 full-body 3D keypoints per person, and (3) depth maps.
In total, MVMP-HMR contains 15 complex scenes, forming a large-scale benchmark for multiview multi-person HMR.
% \paragraph{Task setting.}
% Unlike single-view HMR that typically predicts meshes in camera coordinates, MVMP-HMR requires estimating human meshes in a unified 3D \emph{world} coordinate system.
% This setup is critical for modeling multi-person interactions and scene-level consistency in large scenes.

\begin{table}[t]
\centering
\caption{The statistics of the proposed MVMP-HMR dataset, single-view HMR, and 3D HPE datasets.}
\vspace{-0.15cm}
\setlength{\tabcolsep}{1.0pt}
\renewcommand{\arraystretch}{1.10}
\scriptsize
\begin{tabular}{@{}
>{\centering\arraybackslash}m{0.140\columnwidth}
>{\raggedright\arraybackslash}m{0.170\columnwidth}
>{\centering\arraybackslash}m{0.080\columnwidth}
>{\centering\arraybackslash}m{0.070\columnwidth}
>{\centering\arraybackslash}m{0.085\columnwidth}
>{\centering\arraybackslash}m{0.100\columnwidth}
>{\centering\arraybackslash}m{0.070\columnwidth}
>{\centering\arraybackslash}m{0.195\columnwidth}
@{}}
\toprule
\textbf{Task} & \textbf{Dataset} & \textbf{Area} & \textbf{Scene} & \textbf{Sub.} & \textbf{Occ.} & \textbf{Views} & \makecell[c]{\textbf{GT}\\\textbf{Format}} \\
\midrule

\multirowcell{4}{Single-View\\HMR}
& GTA-Human & - & - & 1 & Simple & 1 & \textbf{SMPL}, J3D \\
& EHF       & - & - & 1 & Simple & 1 & \textbf{SMPLX}, J3D \\
& AGORA     & - & - & 5$\sim$15 & Medium & 1 & \textbf{SMPLX}, Mask \\
& BEDLAM    & - & - & 1$\sim$10 & Medium & 1 & \textbf{SMPLX} \\
\midrule

\multirowcell{3}{Multi-View\\HMR}
& HuMMan & 9$m^{2}$ & 1 & 1 & Simple & 10 & \makecell[c]{\textbf{SMPLX}, J3D,\\Mask} \\
& Hi4D   & 6$m^{2}$ & 1 & 2 & Simple & 53 & \textbf{SMPL}, Mask \\
\midrule

\multirowcell{4}{3D HPE}
& Human3.6M    & 12$m^{2}$ & 7 & 1 & Simple & 4  & \makecell[c]{\textbf{SMPL}, J3D,\\Depth} \\
& 3DPW         & - & - & 1$\sim$2 & Simple & 1  & \textbf{SMPL} \\
& CMU Panoptic & 22$m^{2}$ & 1 & 3$\sim$8 & Medium & 65 & J3D, Depth \\
\midrule

\multirowcell{2}{MVMP-\\HMR}
& Ours & 900$m^{2}$ & 15 & 10$\sim$30 & Severe & 50
& \makecell[c]{\textbf{SMPLX}, J3D,\\Mask, Depth} \\
\bottomrule
\end{tabular}
\vspace{-0.30cm}
\label{tab:datasets}
\end{table}

\textbf{Dataset Annotation.}
Since GTA-V APIs do not provide ground-truth 3D mesh annotations, we generate supervision via a multi-view consistency--driven annotation pipeline that produces both parametric SMPL-X labels and instance masks.
Specifically, for each frame, we first apply a multi-person whole-body HMR method independently to each camera view to obtain candidate SMPL-X predictions in camera coordinates.
We then select the most reliable prediction for each person by minimizing the reprojection error between GTA-V 2D keypoints and the 2D joints projected from the predicted SMPL-X meshes.
The selected SMPL-X annotations are subsequently transformed into a unified world coordinate system by estimating a rigid transformation from shared 3D joint correspondences, yielding world-space parameters \(R_{\text{wld}}^{gt} = R \cdot R_{\text{cam}}\) and \(T_{\text{wld}}^{gt} = T_{\text{cam}} + T\).
In parallel, we generate instance-level human masks using an open-vocabulary segmentation framework based on SAM3\cite{carion2026sam}.
Thus, MVMP-HMR provides consistent world-coordinate SMPL-X annotations together with reliable instance masks, enabling effective supervision for multiview multi-person HMR in large-scale and highly occluded scenes (see Fig.~\ref{fig:Annotation}).

\textbf{Dataset Statistics and Analysis.}
Existing datasets for single-view and multi-view HMR and 3D human pose estimation (HPE) differ substantially from MVMP-HMR in terms of human count, scene scale, viewpoint diversity, and occlusion complexity, as summarized in Table~\ref{tab:datasets}. Overall, existing HMR and HPE datasets are insufficient to represent large-scale, crowded, and highly occluded real-world scenarios. MVMP-HMR addresses these limitations by providing large scenes with dense multi-view coverage, severe occlusions, and rich annotations, making it more suitable for studying multiview multi-person human mesh recovery in realistic settings. We calculated the errors between our SMPL-X annotations (after corrections) and the 3d ground-truth joint points provided by the GTA-V platform and obtained an acceptable error of $4.27cm$, which is relatively small considering the scene size of $30m\times30m\times2m$.
MVMP-HMR also covers diverse environmental conditions generated by the GTA-V engine, which contains a strong bias toward clear or sunny weather (78.12\%), with overcast weather accounting for 12.78\% and adverse weather accounting for 9.09\%.
Its temporal distribution is daytime-oriented, with 79.73\% of samples captured between 6:00 and 18:00 and 20.27\% captured at night.
These statistics reflect practical outdoor scenarios while still preserving variations in illumination, weather, and human activities.

\section{MVMP-HMR Model}
We now introduce our multiview multi-person whole-body human mesh recovery task and model. Given multiview input RGB images $\mathbf{I} = \{I_{1}, I_{2}, \dots, I_{C}\}$ ($C$ is the view number), our model (denoted as $\textbf{f}$), directly predicts a group of ${N}$ centered whole body SMPL-X parameters such as pose ${\theta}\in\mathbb{R}^{N \times 53 \times 3}$, shape $\beta\in\mathbb{R}^{N \times 1 \times 10}$, and expression ${\alpha}\in\mathbb{R}^{N \times 1 \times 10}$, along with their associated 3D spatial translation $\textbf{T}_{wld} \in  \mathbb{R}^{N \times 1 \times 3} $ in the world coordinate system. It outputs expressive 3D human meshes $\textbf{M} = \textbf{SMPL-X} (\textbf{$\theta$}, \textbf{$\beta$}, \textbf{$\alpha$}, \textbf{T}_{wld}) \in \mathbb{R}^{N \times 10475 \times 3}$.
%\begin{align}
%      \{\textbf{M}_{m} + \textbf{T_{wld}_{m}\}_{m \in \{ 1,...,N \}}  = \textbf{f}(\textbf{I})
% \end{align}

%\zq{(Draw a figure showing the pipeline of the paper, especially our multi-view fusion and 3D deformable cov.)}

\begin{figure}[t]
  \centering
  \includegraphics[width=0.98\linewidth]{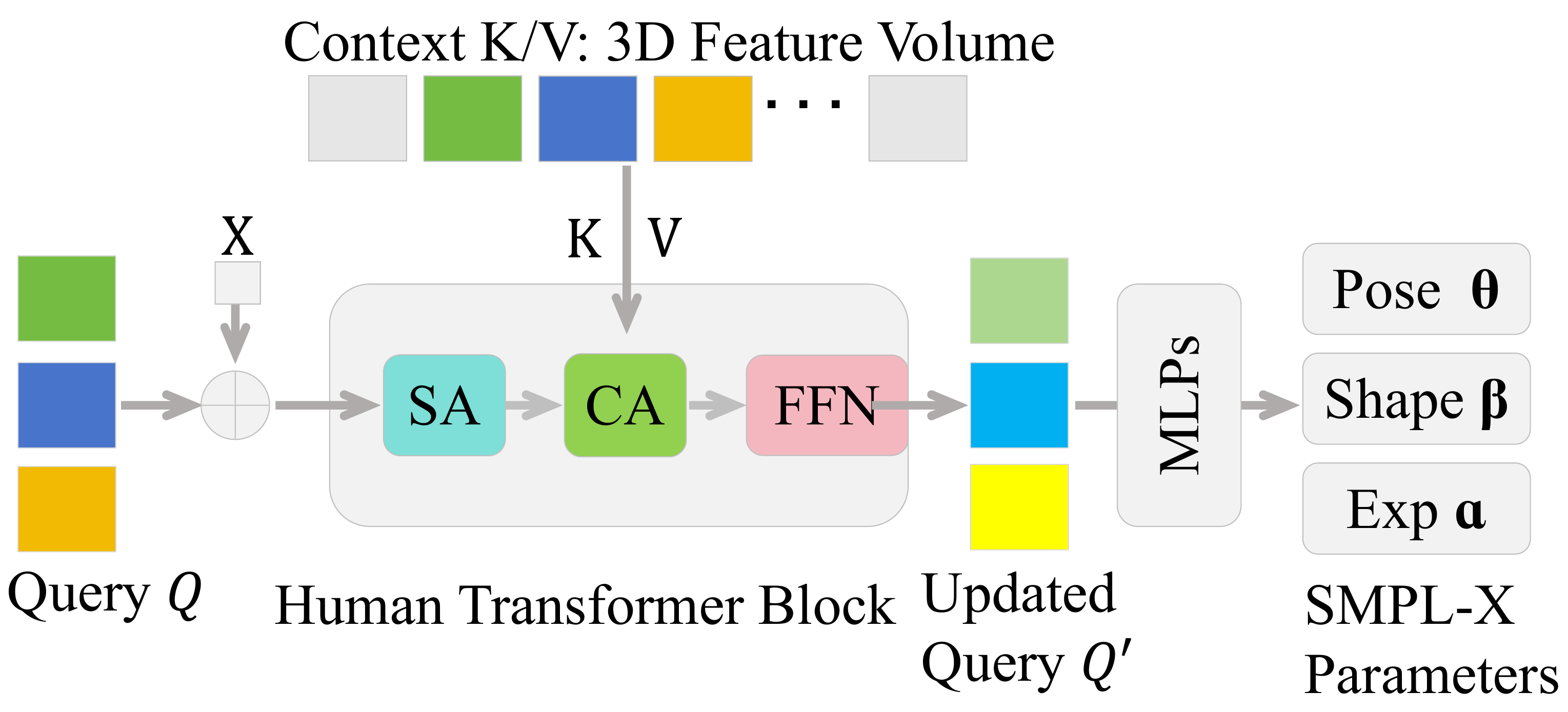}
  \vspace{-0.2cm}
  \caption{Details of the HTB. Human queries are updated by self-attention (SA), cross-attention (CA), and feed-forward layers, then decoded by MLPs for SMPL-X parameter regression.}
  \vspace{-0.4cm}
  \label{fig:hph}
\end{figure}

Unlike single-view HMR, which predicts meshes in camera coordinates and suffers from depth and occlusion ambiguities, MVMP-HMR aims to recover human meshes with absolute 3D locations by leveraging multi-view observations.
As illustrated in Fig.~\ref{fig:pipeline}, our framework consists of three stages: (1) single-view feature extraction using a ViT backbone, (2) multi-view feature projection and fusion to construct a scene-level 3D feature volume, and (3) 3D decoding conditioned on person-specific queries to regress SMPL-X parameters.
This design enables robust mesh recovery for multiple interacting people in large and crowded scenes.

% Compared to single-view human mesh recovery (Single-view HMR), MVMP-HMR task obtains human meshes with absolute locations in 3D world coordinates, rather than relative positions in the camera-view coordinates, because single-view prediction has depth, orientation, pose, and occlusion ambiguities. Thus, MVMP-HMR utilizes multiple views for better multi-view fusion and multi-person mesh recovery to deal with these ambiguities and severe occlusions in practical applications. We require the multiple cameras to be calibrated and synchronized in the setting. As in Figure \ref{fig:pipeline}, the proposed MVMP-HMR model consists of three modules: Single-view Feature Extraction, Multi-view Feature Projection and Fusion, and 3D Decoding, whose details are as below. 

\subsection{Single-view Feature Extraction}
Our MVMP-HMR framework employs the Vision Transformer-Large (ViT-L) \cite{vit} architecture as the single-view feature extractor backbone: $F_{i}= \textbf{ViT-L}(I_{i})_{i \in \{1,...,C\}}$,
% \begin{align}
%     F_{i}= \textbf{ViT-L}(I_{i})_{i \in \{1,...,C\}},    
% \end{align}
where $i$ denotes the view id, $F_{i}$ denotes the feature map of view $I_{i}$, and $C$ is the number of views.
To validate backbone selection, we conduct comprehensive experiments comparing various transformer-based architectures in Sec.~\ref{section:ablation}.%\yt{Ablation study on Backbone}. 
The ViT-L model demonstrates superior performance in capturing global contextual features critical for multi-view fusion. Thus, we use ViT-L as the feature extractor.
In parallel with the ViT-L backbone, an HRNet \cite{hrnet} is employed to predict 2D pose heatmaps $H_i$. After single-view feature extraction, the resulting feature maps ${F_i}$ and heatmaps ${H_i}$ from all views are passed to the fusion stage.
\subsection{Multi-view Feature Projection and Fusion}
%\vspace{-0.2cm}
The extracted single-view features are projected to a constructed 3D volume for multiview feature fusion.
The constructed 3D volume size is $300\times300\times20$, each voxel dimension representing $100\mathrm{mm}$ in the 3D world. So the volume's spatial dimensions are $30\mathrm{m}\times30\mathrm{m}\times2\mathrm{m}$ in the real world.
%Given camera calibration parameters (intrinsic $K$ and extrinsic ${[R\mid t]}$ matrices) provided in our multiview HMR dataset, we establish a unified world coordinate system for 3D volume construction. The scene volume is discretized into 300 grids in length and width, and 20 grids in height. The volume's spatial dimensions are $30\mathrm{m}\times30\mathrm{m}\times2\mathrm{m}$, where each voxel dimension represents $100\mathrm{mm}$ in physical 3D world. This configuration balances the model's performance with computational efficiency.
In the feature projection, we employ perspective geometries to map each 3D voxel coordinate $\mathbf{p}_{w} = (x, y, z)$ to 2D image coordinates of multiple views: $\mathbf{p}_c^{(i)} = \mathbf{K}^{(i)}[\mathbf{R}^{(i)}|\mathbf{t}^{(i)}]\mathbf{p}_w$,
% \begin{equation}
% \mathbf{p}_c^{(i)} = \mathbf{K}^{(i)}[\mathbf{R}^{(i)}|\mathbf{t}^{(i)}]\mathbf{p}_w
% \end{equation}
where intrinsic $\mathbf{K}$ and extrinsic ${[\mathbf{R}\mid \mathbf{t}]}$ matrices are provided in the MVMP-HMR dataset, and $i$ denotes the camera view index. 
We project each view's feature map $F_{i}$ into a 3D volume through this perspective-aware coordinate projection, and each view's 3D feature volume is denoted as $\mathbf{F}_{i}$. 
%Specifically, for each view, features $F_{i}$ are sampled at projected 3D coordinates using bilinear interpolation to construct a single-view feature volume $\mathbf{F}_{i}$. 
Then, we fuse the projected multi-view feature volumes via a mean operation, and the fusion result is denoted as $\mathbf{F_{fuse}}$.
2D heatmaps ${H_{i}}$ are projected into a 3D volume, then fed into a modified RootNet \cite{Voxelpose} to generate 3D probability heatmaps $\mathbf{H}$ (encoding pelvis joint likelihoods in world coordinates). Fusion of these heatmaps yields the coarse 3D grid location $\mathbf{P_{3d}}$ of the primary (pelvis) joint.

% \begin{align}
% & \3DDeformAttn\left(\left\{f_{\hat{v}}\right\}_{\hat{\boldsymbol{v}}=1}^{V}, \boldsymbol{p}, v\right) \\
% = & \sum_{m=1}^{M} \boldsymbol{W}_{m} \sum_{v^{\prime}=1}^{V} \sum_{k=1}^{K} w_{m k v^{\prime}} W_{m}^{\prime} f\left(\boldsymbol{p}+\Delta \boldsymbol{p}_{m k v^{\prime}}\right)
% \end{align}

% Later, the projected features are fused via a XXXX module. 

% \begin{wrapfigure}{r}{0.5\textwidth}
%   \centering
%   \vspace{-0.4cm}
%   \includegraphics[width=1.0\linewidth, height=0.5\linewidth]{figures/HTB.pdf}
%   \caption{The details of the HTB: human queries are updated first via the self-attention layer (SA), the cross-attention layer (CA) integrated with flattened 3D features, and the FeedForward (FFN) layer, and then decoded via MLPs for SMPL-X parameter regression.}
%   \label{fig:hph}
%   \vspace{-0.3cm}
% \end{wrapfigure}

\subsection{3D Decoding}
The fused 3D feature volume $\mathbf{F_{fuse}}$ is decoded with a Human Transformer Block (HTB) to regress the SMPL-X parameters in the 3D world.
%After the above process, the pipeline generates two distinct volumetric representations: a 3D feature volume $\mathbf{F_{final}}$ and a 3D heatmap $\mathbf{H}$. Pelvis localization follows the anatomical prior that this joint serves as the root node in human kinematic chains. 
For each detected human $n\in \{1,..., N\}$ in the 3D heatmap $\mathbf{H}$, we use pelvis joints to sample human features $q$ from $\mathbf{F_{fuse}}$. Then we combine $q$ with $X$ to construct human queries (denoted as $Q$), and $X$ denotes the mean SMPL-X model parameters. Besides, the 3D feature volume $\mathbf{F_{fuse}}$ is flattened as one-dimensional vectors as Keys and Values. Then we input Queries, Keys, and Values into our HTB for SMPL-X parameter regression.
Fig. \ref{fig:hph} shows the details of the Human Transformer Block.
%\yt{ , comprising the self-attention layer (\textbf{SA}), the cross-attention layer (\textbf{CA}), and the FeedForward layer (\textbf{FFN}).}
The full flattened vectors are used as cross-attention keys $K$ and values $V$.
The human queries $Q$ are updated with a stack of $D$ HTB ($D=2$ in practice). 
Then, three MLPs are introduced to regress each human's SMPL-X parameters $\theta$, $\beta$, and $\alpha$ with the updated human queries $Q'$. 

%After these MLPs layers, we can get each human's SMPL-X parameters($\theta$, $\beta$, $\alpha$). 
Human queries $Q$ are also fed into a 3D offset prediction net to estimate the offset $\Delta_{3d}$ of humans. 
Combining the primary joint location $\mathbf{P_{3d}}$ in the 3D heatmap and $\Delta_{3d}$, we can get the final location of the human's primary joint, denoted as translation $\textbf{T}_{wld}=\mathbf{P_{3d}}+\Delta_{3d}$.
%(\zq{Confusion. $T$ is already used in the dataset area.}). 
Finally, we input the SMPL-X parameters and the translation $\textbf{T}_{wld}$ to the SMPL-X layer \cite{EHF} for acquiring humans' mesh vertices and joints locations in world and camera view coordinates.
%\zq{[xxx]}\yt{This is modified from SMPL-X\cite{EHF} for constructing the human meshes, acquiring humans' mesh vertices and joints location in world coordinates and multiview image planes.}

\subsection{Training Loss}
Overall, we adopt five losses to train the proposed MVMP-HMR model.
The first three types of losses are similar to those in the prior work \cite{baradel2024multi}: the \textbf{detection loss} for localizing the human queries, the \textbf{SMPL-X parameter regression loss}, and the \textbf{mesh loss} for supervising 3D joints and vertices coordinate regression in human mesh format. Besides, since our task is in the 3D coordinates system, with orientation and pose ambiguities under the occluded scenes, we propose two novel losses to further enhance the model performance: the \textbf{orientation loss} for better orientation prediction instead of the direct SMPL-X parameters predictions, and the \textbf{3D joint density loss} supervising the predicted 3D joints from the human meshes in 3D density format instead of direct 3D joint coordinate regressing. The details of each loss are as follows:

\textbf{Detection loss}. With the help of the heatmap prediction branch \textbf{HRNet} \cite{hrnet}, we can get the 3D heatmap $\mathbf{H}$ of the primary joint of each human in the scene. Then we construct a 3D volume to present the occupancy of people as $\hat{\textbf{H}}$ with GT joints location. We also obtained the 3D offset $\Delta_{3d}$ in the grid to get a more refined coordinate. So we have the detection loss $\mathcal{L}_{D}$ as follows: 
$\mathcal{L}_{D} = ||\textbf{H} - \hat{\textbf{H}}||_{2} +|\Delta_{3d} - \hat{\Delta}_{3d}|.$
% \begin{align}
% \mathcal{L}_{D} = ||\textbf{H} - \hat{\textbf{H}}||_{2} +|\Delta_{3d} - \hat{\Delta}_{3d}|.
% \end{align}
where $\hat{\textbf{H}}$ and $\hat{\Delta}_{3d}$ are the ground truth 3D heatmap and location offset of the joints, respectively.

\begin{figure}[t]
  \centering
  \includegraphics[width=0.5\linewidth]{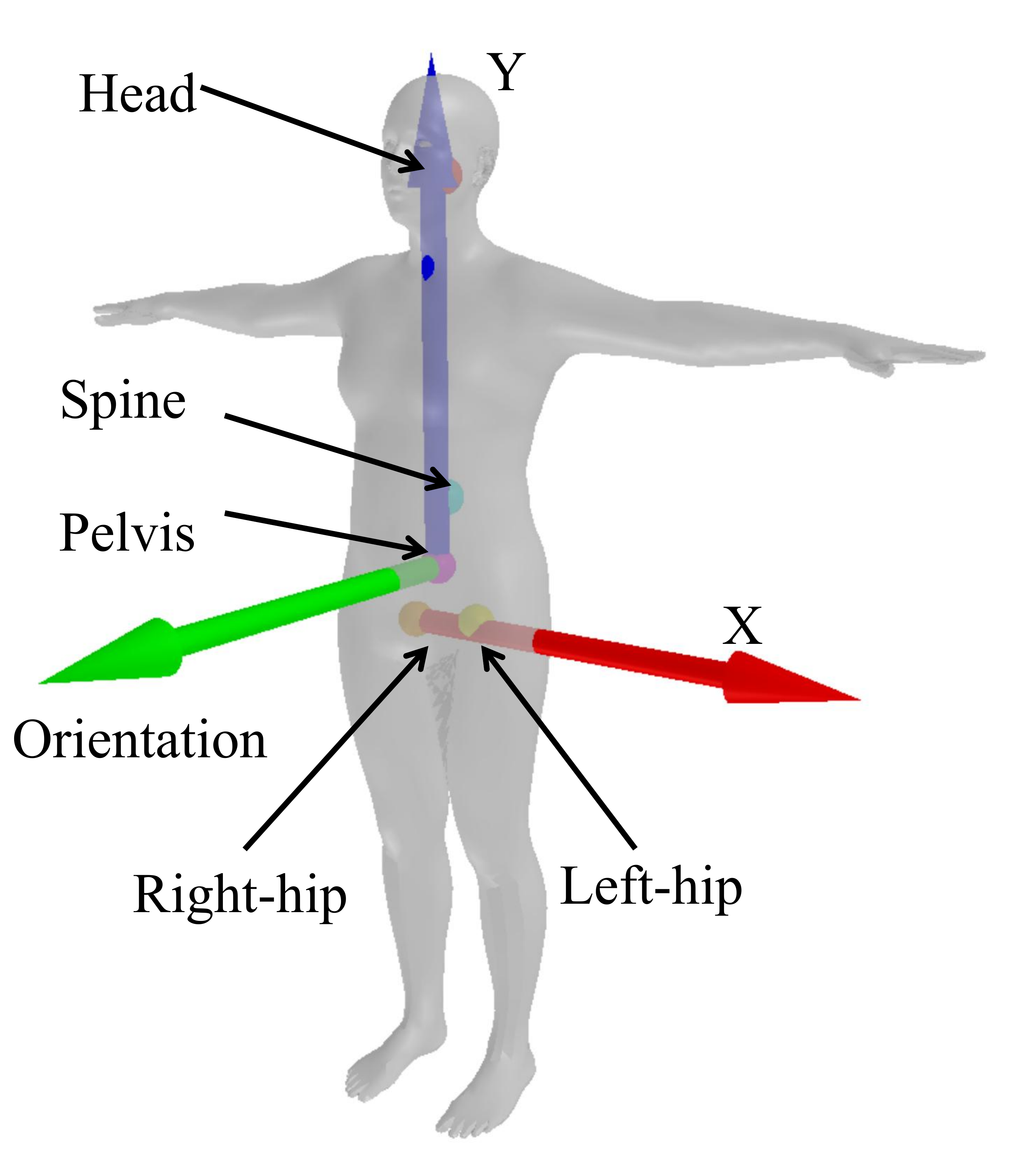}
  \vspace{-0.3cm}
  \caption{The orientation (green) defined from human joints.}
  \vspace{-0.4cm}
  \label{fig:orientation}
\end{figure}

\textbf{Parameter regression loss}. All SMPL-X parameters predicted by the model are computed with \( L_{1}\) regression losses. We integrate the body model parameters (pose $\theta$, shape $\beta$, expression $\alpha$) into loss function as follows:
$\mathcal{L}_{p} = |\theta - \hat{\theta}| + |\beta - \hat{\beta}| + |\alpha - \hat{\alpha}|,$
% \begin{align}
% \mathcal{L}_{p} = |\theta - \hat{\theta}| + |\beta - \hat{\beta}| + |\alpha - \hat{\alpha}|,
% \end{align}
where $\hat{\theta}, \hat{\beta}$, and $\hat{\alpha}$ are the GT parameters. 

\textbf{Mesh loss}. After predicting SMPL-X parameters, we can construct human meshes from a SMPL-X layer. Then we extract 3D joints  $J_{3D}$ and vertices $V_{3D}$ from the human meshes and project these 3D points onto the 2D multi-image planes. %We use the camera projection operator as $\pi_{C}$. 
%We represent 3D joints and vertices as $J_{3D}$ and $V_{3D}$, 2D joints and vertices as $\pi_{c}(J_{3D})$ and $\pi_{c}(V_{3D})$. 
The mesh loss supervises the 3D/2D vertices and joints:
% \begin{align}
% \mathcal{L}_{3D} =  |J_{3D} - \hat{J}_{3D}| + |V_{3D} - \hat{V}_{3D}|,
% \mathcal{L}_{2D}\! =   \!|\pi_{i}(J_{3D})\! - \!\pi_{i}(\hat{J}_{3D})|\! + \!|(\pi_{i}(V_{3D})\! - \!\pi_{i}(\hat{V}_{3D})|,
% \end{align}
\begin{equation}
\begin{gathered}
\mathcal{L}_{3D}
=
\left\| J_{3D}-\hat{J}_{3D} \right\|
+
\left\| V_{3D}-\hat{V}_{3D} \right\|, \\
\mathcal{L}_{2D}
=
\left\| \pi_i(J_{3D})-\pi_i(\hat{J}_{3D}) \right\|
+
\left\| \pi_i(V_{3D})-\pi_i(\hat{V}_{3D}) \right\|.
\end{gathered}
\label{eq:mesh_loss}
\end{equation}
where $\hat{J}_{3D}$ and $\hat{V}_{3D}$ are the ground truth 3D joints and vertices, $\pi_{i}$ is the camera projection operator, and $\pi_{i}(\hat{J_{3D}})$ and $\pi_{i}(\hat{V_{3D}})$ refer to the ground truth 2D joints and vertices projected from the 3D ground truth.
And the mesh loss $\mathcal{L}_{mesh}$ combines the two losses:
$\mathcal{L}_{mesh} =  \lambda_{1}\mathcal{L}_{3D} + \frac{1}{C} \sum_{i=1}^{C} \mathcal{L}_{2D}.$
% \begin{align}
% \mathcal{L}_{mesh} =  \lambda_{1}\mathcal{L}_{3D} + \frac{1}{C} \sum_{i=1}^{C} \mathcal{L}_{2D}.
% \end{align}
Loss weight $\lambda_{1}$ adjusts the weight for the two loss terms and we use a fixed value $\lambda_{1}=100$ in all experiments.

\textit{In addition to these losses, we propose two novel losses:}

% \begin{figure}[t]
%   \centering
%   \includegraphics[width=0.5\linewidth,height=0.5\linewidth]{figures/orientation.pdf}
%   \vspace{-0.5cm}
%   \caption{The orientation (green arrow) defined from human joints.}
%   \vspace{-0.4cm}
%   \label{fig:orientation}
% \end{figure}

\textbf{Orientation loss}. 
The global orientation (a low-dimensional vector) in SMPL-X parameters cannot effectively supervise the orientation of the generated human mesh. Thus, we define the orientation of the human mesh through the joint points for better human mesh orientation supervision (see Fig. \ref{fig:orientation}).
%the global orient (just a low-dimensional vector) can't effectively supervise the orientation of the generated human mesh. 
%So besides the loss of parameters, we determine the orientation of the human mesh through the joint points. As we all know, 
Specifically, a human's left hip $\hat{J}_{lhip}$ and right hip $\hat{J}_{rhip}$ can provide the direction of the x-axis, and a human's pelvis $\hat{J}_{pelvis}$ and spine $\hat{J}_{spine}$ can offer the direction of the y-axis. We use the cross product of the x-axis vector and the y-axis vector to obtain the ground truth orientation  $\hat{\mathcal{O}}$ of the human body:
$\hat{\mathcal{O}} = (\hat{J}_{lhip}-\hat{J}_{rhip}) \times (\hat{J}_{spine}-\hat{J}_{pelvis}).$
% \begin{align}
% \hat{\mathcal{O}} = (\hat{J}_{lhip}-\hat{J}_{rhip}) \times (\hat{J}_{spine}-\hat{J}_{pelvis}).
% \end{align}
In this way, we compute the orientation loss $\mathcal{L}_{\mathcal{O}}$ between the prediction joints $\mathcal{O}$ and ground-truth joints $\hat{\mathcal{O}}$ as:
$\mathcal{L}_{\mathcal{O}} = |\mathcal{O} - \hat{\mathcal{O}}|.$

%So we have computed the Orientation loss (denoted $\mathcal{L}_{o}$) between the prediction joints and GT joints.
% \begin{align}
% \mathcal{L}_{\mathcal{O}} = |\mathcal{O} - \hat{\mathcal{O}}|.
% \end{align}

\textbf{3D joint density loss}. 
%The next one is \textbf{Joint Density map Loss}. 
We use 3D Gaussian kernels to generate a density map of 3D joints from GT  $\hat{J}_{3D}$ and prediction $J_{3D}$. Unlike the direct $L_1$ loss of 3D joint locations (as in mesh loss), 
we use mean square error loss (MSE) for the 3D density map regression:
\begin{align}
%\mathcal{L}_{denj3d} = |\mathcal{O} - \hat{\mathcal{O}}|.
\mathcal{L}_{denj3d} = {\|\textbf{Gau}(J_{3D}) - \textbf{Gau}(\hat{J}_{3D})\|}^2_2.
%\zq{update}
\end{align}
where $\textbf{Gau}$ stands for the Gaussian smoothing step, which generates a 3D Gaussian probability map centered around the joint locations.
The 3D joint density loss $\mathcal{L}_{denj3d}$ is conducted elemental-wisely in 3D space and provides stronger supervision for the pose of the human mesh, handling the pose ambiguities better in the MVMP HMR task under occlusions.
%via spatial smoothing (e.g., Gaussian kernels), %yielding continuous and differentiable loss functions. 
%This 3D spatial supervision is stronger than 
%This encourages the model to learn global density patterns rather than isolated point positions, ensuring coherent 3D mesh recovery even under occlusion or extreme postures. 

In total, the whole training loss is: $\mathcal{L} = \mathcal{L}_{D} + \lambda_{2}\mathcal{L}_{P} +\mathcal{L}_{mesh}  +\lambda_{3} \mathcal{L}_{\mathcal{O}} +  \lambda_{4}\mathcal{L}_{denj3d}$.
% \begin{align}
% \mathcal{L} = \mathcal{L}_{D} + \lambda_{2}\mathcal{L}_{P} +\mathcal{L}_{mesh}  +\lambda_{3} \mathcal{L}_{\mathcal{O}} +  \lambda_{4}\mathcal{L}_{denj3d}.
% \end{align}
We set $\lambda_{2}=10$, $\lambda_{3}=5$ and $\lambda_{4}=1$ in our experiments.

% \begin{align}
% L = \lambda_{1} L_{D} + \lambda_{2} L_{P} + \lambda_{3}L_{3D} + \lambda_{4}L_{2D}  +\lambda_{5} L_{o} +  \lambda_{6}L_{denj3d}
% \end{align}
% \begin{align}
% \mathcal{L} = \mathcal{L}_{D} + \lambda_{2}\mathcal{L}_{P} +\mathcal{L}_{mesh}  +\lambda_{3} \mathcal{L}_{\mathcal{O}} +  \lambda_{4}\mathcal{L}_{denj3d}.
% \end{align}
% We set $\lambda_{2}=10$, $\lambda_{3}=5$ and $\lambda_{4}=1$ in our experiments.
%\yt{there is an parameter $\lambda$ in front of $L_p$.}

%We set \lambda_{1}=1.0, \lambda_{2}=1.0,  \lambda_{3}=100.0, \lambda_{4}=1.0, \lambda_{5}=10.0, \lambda_{6}=0.1
%\zq{Is there any loss weight?}

\section{Experiments and Results}

\subsection{Experiment Settings}
\label{Implementation} 

\textbf{Datasets and Implementation}.  
We perform the experiments on 3 datasets: \textit{MVMP-HMR} collected by us, and 2 real datasets \textit{CMU Panoptic} \cite{joo2015panoptic}, and \textit{Human3.6M} \cite{Human3.6m}. In experiments, we divide the 15 scenes in our dataset according to the distribution of people numbers, and the ratio of the training/testing set is 2:1.
We use VIT-L \cite{vit} as our model feature extraction backbone. We pre-train the posenet \cite{hrnet} and rootnet \cite{Voxelpose} for 60 epochs on our dataset for detection. The input images are resized to 1288 x 1288 with zero padding. We adopt Adam as the optimizer with 5e-5 learning rate. The training epoch is 50, and the training is conducted on 2 RTX6000 Ada GPUs, with a batch size of 1.

\textbf{Evaluation Metrics}. 
We evaluate HMR predictions in the 3D world rather than in an individual camera coordinate system.
MPJPE measures the average Euclidean distance between predicted 3D joints and ground-truth 3D joints.
PVE measures the average point-to-point Euclidean distance between predicted mesh vertices and ground-truth mesh vertices in world space.
PA-PVE computes PVE after Procrustes alignment between predicted and ground-truth mesh vertices.
MPJPE and PVE are the main metrics in our task. All keypoints and vertices are obtained from the corresponding SMPL-X parameters, and all reported errors are measured in millimeters.

\begin{table*}[t]
\caption{Result comparison on MVMP-HMR, Human3.6M, and CMU Panoptic. The best results are in \textbf{bold}, and the second-best results are \underline{underlined}.}
\label{table:main_results}
\vspace{-0.2cm}
\centering
\scriptsize
\setlength{\tabcolsep}{4.2pt}
\renewcommand{\arraystretch}{1.15}

\begin{tabular}{l|ccc|ccc|ccc}
\toprule
Dataset
& \multicolumn{3}{c|}{MVMP-HMR}
& \multicolumn{3}{c|}{Human3.6M}
& \multicolumn{3}{c}{CMU Panoptic} \\
Method
& MPJPE$\downarrow$ & PVE$\downarrow$ & PA-PVE$\downarrow$
& MPJPE$\downarrow$ & PVE$\downarrow$ & PA-PVE$\downarrow$
& MPJPE$\downarrow$ & PVE$\downarrow$ & PA-PVE$\downarrow$ \\
\midrule

3DCrowdNet (Dist)   & 221.2 & 284.3 & 72.2  & 443.2 & 456.3 & 186.7 & 135.8 & 130.9 & 69.5 \\
AiOS (Dist)         & 873.6 & 642.4 & 110.5 & 156.8 & 133.4 & 78.9  & 730.6 & 550.9 & 195.8 \\
TokenHMR (Dist)     & 632.3 & 661.3 & 191.5 & 112.4 & 122.5 & 58.9  & 616.3 & 598.0 & 194.4 \\
Multi-HMR (Dist)    & 841.0 & 651.4 & 71.0  & 98.5  & 97.3  & 46.3  & 568.7 & 453.4 & 195.1 \\
Multi-HMR (Avg)     & 752.5 & 753.6 & 61.7  & 110.3 & 99.8  & 52.7  & 546.9 & 509.8 & 220.8 \\
Multi-HMR (Fusion)  & 602.4 & 529.5 & 111.4 & 129.7 & 122.8 & 65.4  & 523.3 & 423.1 & 192.6 \\
HeatFormer          & \underline{185.5} & \underline{148.3} & \underline{83.6}
                           & \textbf{60.3}  & \textbf{65.4}  &  \textbf{31.2}
                           & 385.6          & 376.2     & \underline{125.6} \\ 
\midrule
    VoxelSMPLX (Only)          & 225.4 & 262.0 & 240.6  
                               & 147.5 & 160.3 & 54.5   
                               & \underline{372.1} & \underline{365.5} & 134.2 \\
    VoxelSMPLX (Joint)         & 288.6 & 427.4 & 317.1  
                               & 156.3 & 167.2 & 61.3   
                               & 403.5 & 385.9 & 158.6\\
\midrule
MVMP-HMR (Ours)
& \textbf{177.5} & \textbf{129.2} & \textbf{51.8}
& \underline{93.5}  & \underline{92.1}  & \underline{44.3}
& \textbf{278.6} & \textbf{234.5} & \textbf{95.3} \\
\bottomrule
\end{tabular}

\vspace{-0.2cm}
\end{table*}

\begin{figure*}[t]
  \centering
  \includegraphics[width=0.75\linewidth]{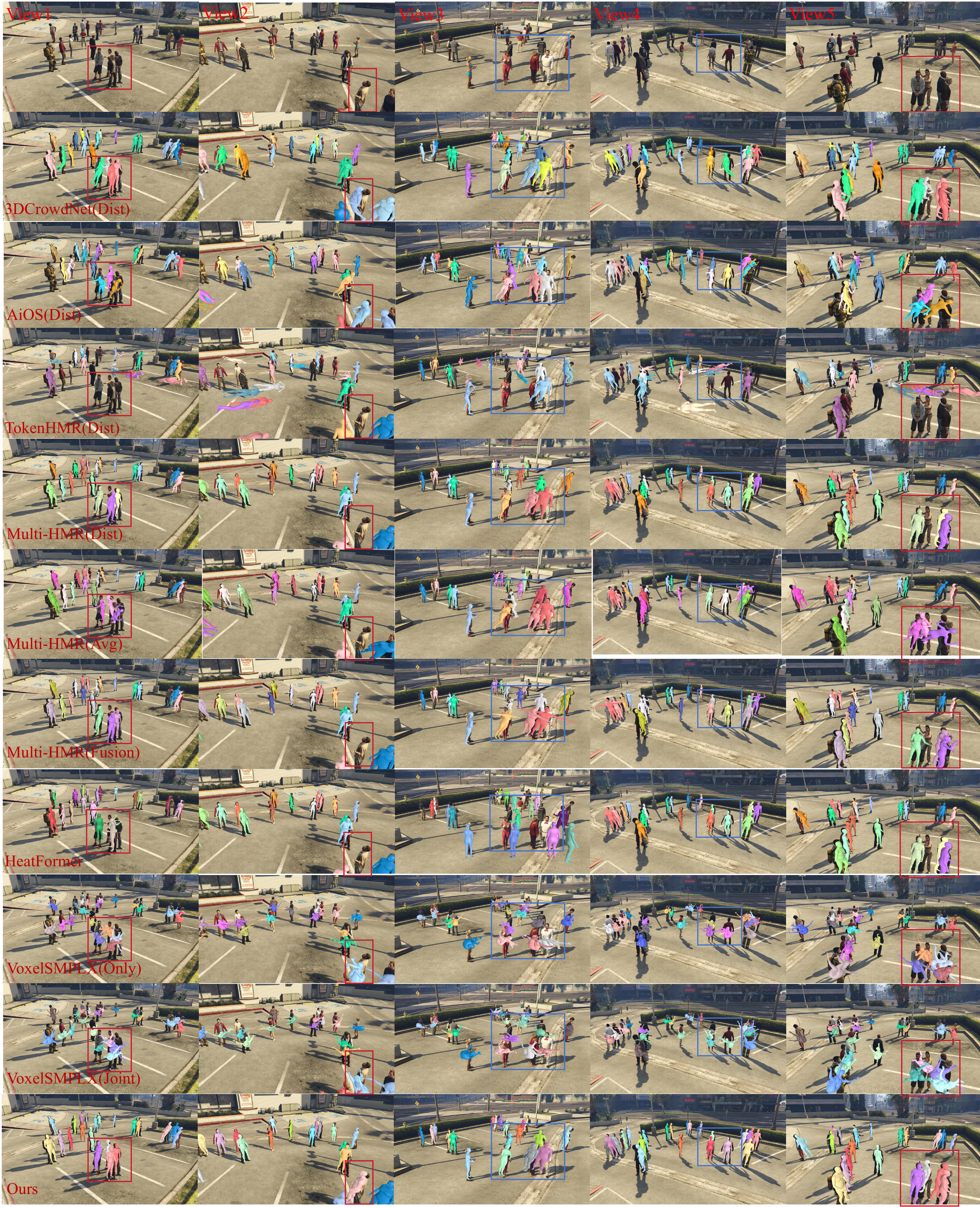}
  \vspace{-0.1cm}
  \caption{The top row is the multiview input, and each subsequent row is the 3D predictions of the methods projected to the view plane. Red boxes indicate that our method can better handle occlusions than comparison methods. Blue boxes indicate our method achieves better posture than the comparisons.
  }
  \vspace{-0.1cm}
  \label{fig:result}
\end{figure*}

\textbf{Comparison methods.}
We compare our MVMP-HMR method with multi-person HMR SOTAs with multiview settings, 3D HPE method for HMR tasks, and a multiview single-person method HeatFormer \cite{matsubara2025heatformer}.
Single-view HMR SOTAs Multi-HMR \cite{baradel2024multi}, 3DCrowdNet \cite{choi2022learning}, AiOS \cite{sun2024aios}, and TokenHMR \cite{dwivedi2024tokenhmr} first conduct predictions of each view, then use a multi-view matching algorithm to match the prediction results of each person under multiple views, and fuse the prediction results of each person in the scene under multiple views into the final result. The fusion strategy includes selecting the closest one as the prediction result based on the distance from the camera (denoted as `Dist'), using an average strategy to fuse the results of each view prediction (denoted as `Avg'), and using a sub-network to predict the weight value corresponding to each view prediction to fuse the final result (denoted as `Fusion').
% We compare our Multiview-HMR method with multi-person HMR SOTAs with multiview settings and 3D HPE method for HMR tasks.
% \textbf{Distance-selected Method.} We compare with single-view HMR SOTAs Multi-HMR \cite{baradel2024multi}, 3DCrowdNet \cite{crowd3d}, AiOS \cite{sun2024aios} and TokenHMR \cite{dwivedi2024tokenhmr} by selecting the closest camera's prediction for each person since a close camera may provide more accurate predictions. These methods are denoted as `{Multi-HMR (Dist)}', `{3DCrowdNet (Dist)}', `AiOS (Dist)' and `TokenHMR (Dist)'.
% \textbf{Fusion-based method.} We also fuse the multiview SMPL-X parameter predictions of Multi-HMR \cite{baradel2024multi} with a learnable subnet, where the multiview SMPL-X predictions are weighted and summed via a self-attention layer, denoted as `{Multi-HMR (Fusion)}'.
We also compare with a multi-view 3D pose estimation method VoxelPose \cite{Voxelpose}.
We sample human queries from the feature volume with the predicted joint locations of VoxelPose \cite{Voxelpose} and then estimate the SMPL-X parameters from the human queries with regression MLPs. There are two variants: use the pretrained VoxelPose and only train the regression MLPs, denoted `{VoxelSMPLX (Only)}'; or jointly train VoxelPose and MLPs, denoted as `{VoxelSMPLX (Joint)}'. We extend the single-person HeatFormer \cite{matsubara2025heatformer} to multi-person scenarios via a top-down framework, decomposing the scene into individual instances for independent SMPL-X regression.

\subsection{Results On MVMP-HMR Dataset}

We evaluate all methods on three benchmarks--MVMP-HMR (ours), Human3.6M, and CMU Panoptic--as summarized in Table~\ref{table:main_results}. We compare against six single-view HMR baselines augmented with multi-view fusion, a 3D HPE method with SMPL-X regression (VoxelSMPLX), and HeatFormer, a recent multi-view transformer designed for single-person mesh recovery.
MVMP-HMR achieves the best performance on the proposed MVMP-HMR benchmark and CMU Panoptic, while remaining competitive on Human3.6M. Existing methods are primarily designed for single-view HMR or 3D pose estimation in relatively simple scenes, limiting their ability to robustly fuse multi-view information or recover accurate meshes in crowded settings. 
Although HeatFormer achieves the best performance on Human3.6M, this is expected since the dataset contains only a single subject per scene and aligns well with HeatFormer's single-person formulation. 
These results further highlight the advantage of our MVMP-HMR model in complex multi-person scenes with significant occlusion and orientation ambiguity.

\begin{figure*}[t]
  \centering
  \includegraphics[width=0.8\linewidth]{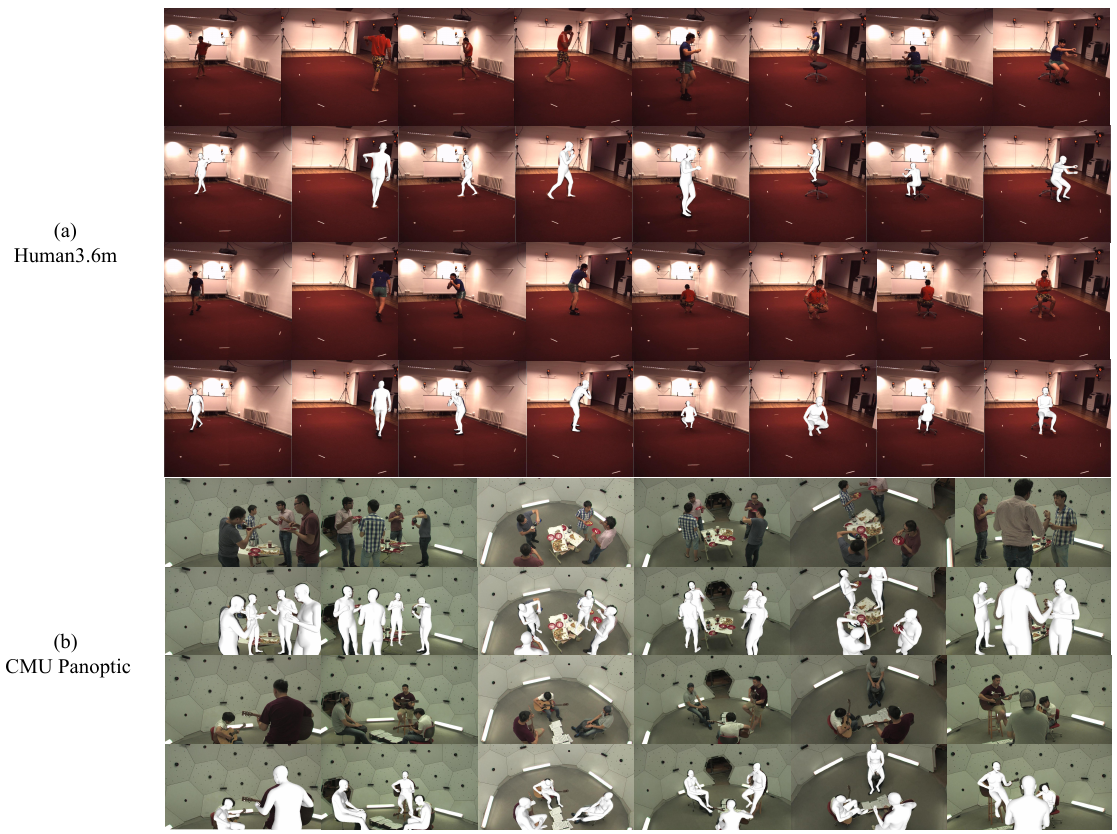}
  \caption{Qualitative results on real-world datasets: (a) Human3.6m, and (b) CMU Panoptic.}
  \label{fig:vis_human3.6m}
\end{figure*}

\textbf{Visualizations}. Fig. \ref{fig:result} shows that our proposed method outperforms all comparison methods on the MVMP-HMR dataset, where the \textit{red boxes} indicate our method can handle occlusions well and estimate meshes more accurately for occluded persons, and the \textit{blue boxes} show our method achieves more natural and realistic human poses, with better limb positioning and alignment.
Besides, Fig. \ref{fig:vis_human3.6m} shows our results on the two real datasets Human3.6M and CMU Panoptic, where our method remains robust and reconstructs accurate meshes for all individuals in real-world scenes.

\begin{table}[t]
\scriptsize
\centering
\caption{Loss term weight ablation study. The first row uses no proposed losses, Rows 2--5 add the orientation loss, and Rows 6--11 add both orientation and 3D joint density losses.}
\vspace{-0.1cm}
\setlength{\tabcolsep}{5pt}
\begin{tabular}{l|ccc}
%@{\hspace{0.02cm}}l@{\hspace{0.12cm}}|c@{\hspace{0.12cm}}c@{\hspace{0.12cm}}|c@{\hspace{0.12cm}}c@{\hspace{0.12cm}}|c@{\hspace{0.12cm}}c@{\hspace{0.02cm}}}
\toprule %\hline
    Loss                     & MPJPE $\downarrow$          & PVE  $\downarrow$           & PA-PVE   $\downarrow$   \\
\midrule %\hline
%\multirow{4}{*}{None-VTR}
    $\mathcal{L}_{D} \!+ \! \lambda_{2}\mathcal{L}_{P} \!+\! \mathcal{L}_{mesh}$       & 217.1&   161.7&  120.4    \\ %\citep{mescheder2019occupancy} 
    % $\ +\lambda_3\mathcal{L}_{\mathcal{O}}$                & 195.1 &  149.2 &  80.7    \\ %\citep{xie2020pix2vox++}
    % $\ +\lambda_{4}\mathcal{L}_{denj3d}$              & 180.7&   132.4&  50.2    \\ 
    $\ +2.0*\mathcal{L}_{\mathcal{O}}$                  & 201.6 & 151.9 & 99.4   \\ 
    $\ +\textbf{5.0}*\mathcal{L}_{\mathcal{O}}$         & 187.9 & 144.8 & 89.0   \\ 
    $\ +10.0*\mathcal{L}_{\mathcal{O}}$                 & 195.1 & 149.2 & 80.7  \\
    $\ +100.0*\mathcal{L}_{\mathcal{O}}$                & 195.0 & 150.0 & 71.8 \\
    $\ +5.0*\mathcal{L}_{\mathcal{O}} +0.1*\mathcal{L}_{denj3d}$                              & 190.2 & 144.5 & 89.4 \\
    $\ +5.0*\mathcal{L}_{\mathcal{O}} +0.2*\mathcal{L}_{denj3d}$                              & 190.7 & 149.7 & 83.3 \\
    $\ +5.0*\mathcal{L}_{\mathcal{O}} +0.5*\mathcal{L}_{denj3d}$                              & 187.6 & 147.4 & 87.9 \\
    $\ +\textbf{5.0}*\mathcal{L}_{\mathcal{O}} +\textbf{1.0}*\mathcal{L}_{denj3d}$ (\textbf{Ours})                              & \textbf{177.5} &  \textbf{129.2} &  \textbf{51.8} \\
    $\ +5.0*\mathcal{L}_{\mathcal{O}} +2.0*\mathcal{L}_{denj3d}$                              & 293.3 & 149.5 & 69.6 \\
    $\ +5.0*\mathcal{L}_{\mathcal{O}} +5.0*\mathcal{L}_{denj3d}$                              & 368.8 & 160.0 & 69.4 \\
    %$\ +$Both (Ours)                  & \textbf{177.5} &  \textbf{129.2} &  \textbf{51.8} \\ % \citep{yang2023long}  
\bottomrule %\hline
\end{tabular}
\vspace{-0.1cm}
\label{table:loss}
\end{table}

% \vspace{-0.25cm}
\subsection{Ablation Study}
\label{section:ablation}

\setcounter{topnumber}{4}
\renewcommand{\topfraction}{0.99}
\renewcommand{\textfraction}{0.01}

\textbf{Loss term weight ablation study.}
We conduct the loss term weight ablations for the proposed orientation loss ($\mathcal{L}_\mathcal{O}$) and 3D joint density loss ($\mathcal{L}_{denj3d}$) in Table \ref{table:loss}. The first row uses the loss usually used in prior work \cite{baradel2024multi}. Row 2-5 add the proposed orientation loss $\mathcal{L}_\mathcal{O}$ with different $\lambda_{3}$ weights, and the performance all improved compared to without it, demonstrating the effectiveness of the $\mathcal{L}_\mathcal{O}$ loss. $\lambda_{3}=5.0$ achieves the best results, and we use it as the loss weight of $\mathcal{L}_\mathcal{O}$ in the experiments. 
Row 6-11 further add the proposed 3D joint density loss $\mathcal{L}_{denj3d}$ in the model training. $\lambda_{3}=5.0, \lambda_{4}=1.0$ achieves the best results. When $\lambda_{4}$ is too large, the 3D joint density loss may decrease the human mesh prediction performance because $\mathcal{L}_{denj3d}$ might be too strong.

\begin{table}[t]
\small
\centering
\caption{Feature fusion ablation study.}
\begin{tabular}{l|ccc}
\toprule
Fusion & MPJPE$\downarrow$ & PVE$\downarrow$ & PA-PVE$\downarrow$ \\
\midrule
Deform. & 261.3 & 207.2 & 80.6 \\
Max     & 245.2 & 193.5 & 74.8 \\
Mean    & \textbf{177.5} & \textbf{129.2} & \textbf{51.8} \\
\bottomrule
\end{tabular}
\vspace{-0.2cm}
\label{table:fusion_ablation}
\end{table}

% \begin{table}[t]
%     \centering
%     \begin{tabular}{c|c}
%          &  \\
%          & 
%     \end{tabular}
%     \caption{Caption}
%     \label{tab:placeholder}
% \end{table}
% %\begin{minipage}{\linewidth}
% \centering
% \captionof{table}{Feature fusion ablation study with MPJPE, PVE, and PA-PVE metrics.}
% \label{table:fusion_ablation}
% \vspace{-0.05cm}
% \scriptsize
% \setlength{\tabcolsep}{3pt}
% \renewcommand{\arraystretch}{0.92}
% \begin{tabular}{l|ccc}
% \toprule
% Fusion & MPJPE$\downarrow$ & PVE$\downarrow$ & PA-PVE$\downarrow$ \\
% \midrule
% Deform. & 261.3 & 207.2 & 80.6 \\
% Max     & 245.2 & 193.5 & 74.8 \\
% Mean    & \textbf{177.5} & \textbf{129.2} & \textbf{51.8} \\
% \bottomrule
% \end{tabular}
% \vspace{0.05cm}
% %\end{minipage}
% \end{table}

\textbf{Feature fusion method ablation study.}
We also perform ablation studies on the feature fusion method, using three different methods: Deformable attention, Max, and Mean. As shown in Table~\ref{table:fusion_ablation}, the performance using the mean operation to fuse multi-view features achieves marginally superior performance than using deformable attention or max. The possible reason is that the mean method is simple and efficient, suitable for global information fusion, but the max method is suitable for highlighting key features, but is susceptible to noise interference. And the deformable attention has a high computational overhead. In our setting, the mean operation is better for aggregating multi-view features. Thus, in our experiments, we use `Mean' as the feature fusion method.

% \begin{center}
% \begin{minipage}{\linewidth}
% \centering
% \captionof{table}{Primary joint ablation study with MPJPE, PVE, and PA-PVE metrics.}
% \label{table:joint_ablation}
% \vspace{-0.05cm}
% \scriptsize
% \setlength{\tabcolsep}{3pt}
% \renewcommand{\arraystretch}{0.92}
% \begin{tabular}{l|ccc}
% \toprule
% Joint & MPJPE$\downarrow$ & PVE$\downarrow$ & PA-PVE$\downarrow$ \\
% \midrule
% Head   & 280.6 & 172.3 & 68.2 \\
% Spine  & 190.2 & 146.9 & 86.1 \\
% Pelvis & \textbf{177.5} & \textbf{129.2} & \textbf{51.8} \\
% \bottomrule
% \end{tabular}
% \vspace{0.10cm}
% \end{minipage}
% \end{center}

\begin{table}[t]
\small
\centering
\caption{Primary joint ablation study.}
\vspace{-0.1cm}
\setlength{\tabcolsep}{6pt}
\renewcommand{\arraystretch}{0.95}
\begin{tabular}{l|ccc}
\toprule
Joint & MPJPE$\downarrow$ & PVE$\downarrow$ & PA-PVE$\downarrow$ \\
\midrule
Head   & 280.6 & 172.3 & 68.2 \\
Spine  & 190.2 & 146.9 & 86.1 \\
Pelvis & \textbf{177.5} & \textbf{129.2} & \textbf{51.8} \\
\bottomrule
\end{tabular}
\vspace{-0.2cm}
\label{table:joint_ablation}
\end{table}

% \textbf{Primary joint selection ablation study.}
% To determine the optimal primary joint for our model, we conducted an ablation study comparing three different primary joints: the pelvis, head, and spine. As shown in Table~\ref{table:joint_ablation}, the results show that the use of the pelvis for localisation produces marginally better performance. This can be attributed to the pelvis's stability across various viewpoints and its central location, which allows for more complete human body information to be captured in the model's queries. Consequently, we chose the pelvis as the primary joint for all subsequent experiments.

\textbf{Primary joint selection ablation study.}
Table~\ref{table:joint_ablation} compares pelvis, head, and spine as the primary joint for query extraction.
The pelvis achieves the best results because it is more stable across viewpoints and provides a central cue for whole-body feature sampling.
Thus, we use the pelvis as the primary joint in all subsequent experiments.

% \begin{center}
% \begin{minipage}{\linewidth}
% \centering
% \captionof{table}{Backbone ablation study with MPJPE, PVE, and PA-PVE metrics.}
% \label{table:backbone_ablation}
% \vspace{-0.05cm}
% \scriptsize
% \setlength{\tabcolsep}{3pt}
% \renewcommand{\arraystretch}{0.92}
% \begin{tabular}{l|ccc}
% \toprule
% Backbone & MPJPE$\downarrow$ & PVE$\downarrow$ & PA-PVE$\downarrow$ \\
% \midrule
% ViT-S & 201.6 & 157.8 & 64.9 \\
% ViT-B & 185.7 & 141.8 & 61.6 \\
% ViT-L & \textbf{177.5} & \textbf{129.2} & \textbf{51.8} \\
% \bottomrule
% \end{tabular}
% \vspace{0.05cm}
% \end{minipage}
% \end{center}

\textbf{Feature extraction backbone ablation study.}
We compare three ViT backbones with different model sizes: ViT-S, ViT-B, and ViT-L \cite{vit}.
As shown in Table~\ref{table:backbone_ablation}, ViT-L achieves the best results, indicating that a stronger feature extractor is beneficial for global context modeling and multiview fusion.
Therefore, ViT-L is adopted as the default backbone.

\begin{table}[t]
\small
\centering
\vspace{-0.08cm}
\caption{Backbone model ablation study.}
\setlength{\tabcolsep}{6pt}
\renewcommand{\arraystretch}{0.95}
\begin{tabular}{l|ccc}
\toprule
Backbone & MPJPE$\downarrow$ & PVE$\downarrow$ & PA-PVE$\downarrow$ \\
\midrule
ViT-S & 201.6 & 157.8 & 64.9 \\
ViT-B & 185.7 & 141.8 & 61.6 \\
ViT-L & \textbf{177.5} & \textbf{129.2} & \textbf{51.8} \\
\bottomrule
\end{tabular}
\vspace{-0.2cm}
\label{table:backbone_ablation}
\end{table}

% \begin{center}
% \begin{minipage}{\linewidth}
% \centering
% \captionof{table}{Testing view number ablation study with MPJPE, PVE, and PA-PVE metrics.}
% \label{table:view_ablation}
% \vspace{-0.05cm}
% \scriptsize
% \setlength{\tabcolsep}{3pt}
% \renewcommand{\arraystretch}{0.92}
% \begin{tabular}{l|ccc}
% \toprule
% Views & MPJPE$\downarrow$ & PVE$\downarrow$ & PA-PVE$\downarrow$ \\
% \midrule
% 3 & 193.6 & 137.6 & 50.9 \\
% 5 & 177.5 & 129.2 & 51.8 \\
% 7 & 171.0 & 125.2 & 48.2 \\
% 9 & \textbf{168.1} & \textbf{122.0} & \textbf{47.9} \\
% \bottomrule
% \end{tabular}
% \vspace{0.05cm}
% \end{minipage}
% \end{center}

\begin{table}[t]
\small
\centering
\caption{Testing view number ablation study: trained with 5 views and tested on 3-9 views.}
\setlength{\tabcolsep}{6pt}
\renewcommand{\arraystretch}{0.95}
\begin{tabular}{c|ccc}
\toprule
Views & MPJPE$\downarrow$ & PVE$\downarrow$ & PA-PVE$\downarrow$ \\
\midrule
3 & 193.6 & 137.6 & 50.9 \\
5 & 177.5 & 129.2 & 51.8 \\
7 & 171.0 & 125.2 & 48.2 \\
9 & \textbf{168.1} & \textbf{122.0} & \textbf{47.9} \\
\bottomrule
\end{tabular}
\vspace{-0.2cm}
\label{table:view_ablation}
\end{table}

\textbf{Testing view number ablation study.}
We further evaluate the robustness of MVMP-HMR to different numbers of testing cameras.
The model is trained with 5 camera views and tested with 3, 5, 7, and 9 views, as shown in Table~\ref{table:view_ablation}.
Performance improves as more camera views are provided because additional observations offer richer cues for resolving occlusion and depth ambiguity.
The performance gap remains moderate when reducing the number of testing views, demonstrating that our model generalizes well to different camera configurations.

\begin{table}[t]
\small
\centering
\caption{The model parameters and inference time compared to HMR and HPE SOTAs.}
\begin{tabular}{l|cc@{\hspace{0.02cm}}}
%@{\hspace{0.02cm}}l@{\hspace{0.12cm}}|c@{\hspace{0.12cm}}c@{\hspace{0.12cm}}|c@{\hspace{0.12cm}}c@{\hspace{0.12cm}}|c@{\hspace{0.12cm}}c@{\hspace{0.02cm}}}
\toprule %\hline
    Method     & Parameters (MB) $\downarrow$ & Inference (s)  $\downarrow$  \\
\midrule %\hline
    3DCrowdNet (Dist) \cite{choi2022learning}              & 931.92 & 1.12 \\
    AiOS (Dist) \cite{sun2024aios}                 & 1122.28& 0.97  \\
    TokenHMR (Dist) \cite{dwivedi2024tokenhmr}     & 2598.57 & 2.44  \\
    Multi-HMR (Dist) \cite{baradel2024multi}           & 1210.17 & 2.33 \\ %\cite{mescheder2019occupancy} 
    Multi-HMR (Avg) \cite{baradel2024multi}           & 1210.17 & 2.33 \\ %\cite{mescheder2019occupancy} 
    Multi-HMR (Fusion) \cite{baradel2024multi}         & 1331.19 & 2.53 \\ % \cite{yang2023long}  
    HeatFormer \cite{matsubara2025heatformer}     & 3260.58 & 1.78    \\
    VoxelSMPLX (Only) \cite{Voxelpose}         & 404.45 & 1.00 \\ % 
    VoxelSMPLX (Joint) \cite{Voxelpose}         & 404.45 & 1.00 \\ % 
\midrule %\hline
    MVMP-HMR (Ours)   & 1380.28 &  1.59 \\
\bottomrule %\hline
\end{tabular}
\vspace{-0.4cm}
\label{table:eff}
\end{table}

\textbf{Model complexity and inference time analysis.}
In addition to the results displayed in the dataset compared with other methods, we also compare model parameters and inference speed in Table \ref{table:eff}. Our model parameters only count the model parameters during testing. Inference time is measured by running the model on 100 input samples and averaging the runtime. From our model framework, it can be seen that the 3D voxel features constructed from multi-view feature projections and fusion, as well as the subsequent network processing, are very resource-intensive. However, our model's parameters and inference speed achieve a moderate result compared to existing HMR and multi-view HPE methods. Although the HPE method has a simpler network architecture, resulting in lower estimated model parameters and inference speed than ours, the HPE method can't achieve good results on our MVMP-HMR dataset. Single-view HMR does not involve the fusion of multi-view features, so its model parameter count is smaller than ours. Additionally, the efficiency of detecting directly on 3D voxel features is higher than that of multi-view matching, leading to shorter inference times for our method.

\begin{table}[t]
\scriptsize
\centering
\caption{Cross-scene generalization: with the help of the proposed synthetic dataset, the cross-scene generalization ability is improved for both HeatFormer and ours.}
%\textit{Top}: zero-shot cross-domain transfer (trained on one dataset, tested on another without fine-tuning). \textit{Bottom}: synthetic-to-real augmentation (Panoptic training augmented with MVMP-HMR). Values in parentheses denote change relative to the in-domain baseline.}
\vspace{-0.1cm}
\resizebox{\linewidth}{!}{%
\begin{tabular}{@{\hspace{0.0cm}}l@{\hspace{0.01cm}}l@{\hspace{0.01cm}}c@{\hspace{0.01cm}}c@{\hspace{0.01cm}}c@{\hspace{0.0cm}}}
\toprule
Train $\rightarrow$ Test & Method & MPJPE $\downarrow$ & PVE $\downarrow$ & PA-PVE $\downarrow$ \\
\midrule
H3.6M $\rightarrow$ H3.6M        & HeatFormer & 60.3  & 65.4  & 31.2 \\
MVMP-HMR $\rightarrow$ H3.6M     & HeatFormer & 228.6 {\scriptsize(+168.3)} & 247.5 {\scriptsize(+182.1)} & 157.6 {\scriptsize(+126.4)} \\
\cmidrule{3-5}
H3.6M $\rightarrow$ H3.6M        & Ours       & 93.5  & 92.1  & 44.3 \\
MVMP-HMR $\rightarrow$ H3.6M     & Ours       & 139.8 {\scriptsize(+46.3)}  & 135.5 {\scriptsize(+43.4)}  & 91.0 {\scriptsize(+46.7)}  \\
\midrule
Panoptic $\rightarrow$ Panoptic            & HeatFormer & 385.6 & 376.2 & 125.6 \\
Panoptic+MVMP-HMR $\rightarrow$ Panoptic  & HeatFormer & 376.8 {\scriptsize($-$8.8)}  & 369.1 {\scriptsize($-$7.1)}  & 126.4 {\scriptsize(+0.8)}  \\
\cmidrule{3-5}
Panoptic $\rightarrow$ Panoptic            & Ours       & 278.6 & 234.5 & 95.3 \\
Panoptic+MVMP-HMR $\rightarrow$ Panoptic  & Ours       & \textbf{257.8} {\scriptsize($\mathbf{-20.8}$)} & \textbf{218.3} {\scriptsize($\mathbf{-16.2}$)} & 98.2 {\scriptsize(+2.9)} \\
\bottomrule
\end{tabular}
}
\vspace{-0.3cm}
\label{tab:crossscene}
\end{table}

% \vspace{-0.5cm}
\subsection{Cross-Scene Generalization}
\label{section:crossscene}
To evaluate the cross-scene generalization of our proposed framework, we conduct two complementary experiments in Table~\ref{tab:crossscene}: 
(1) \textbf{Zero-shot cross-domain transfer}, where models trained on one dataset are directly tested on another without fine-tuning: We train both our method and HeatFormer~\cite{matsubara2025heatformer} on MVMP-HMR and evaluate directly on Human3.6M without any fine-tuning.
As shown in Table~\ref{tab:crossscene}, a noticeable domain gap exists for both methods due to the synthetic-to-real appearance shift.
Nevertheless, our method (MPJPE: 139.8) outperforms HeatFormer (MPJPE: 228.6) by a large margin under the same zero-shot setting, demonstrating that our geometry-aware world-coordinate 3D fusion provides stronger cross-scene robustness.
(2) \textbf{Synthetic-to-real augmentation}, where MVMP-HMR synthetic data is used as additional training data to improve performance on real datasets: To further assess whether MVMP-HMR benefits real-world performance, we augment CMU Panoptic training with MVMP-HMR synthetic data and evaluate on the Panoptic test set.
As shown in Table~\ref{tab:crossscene}, adding MVMP-HMR consistently improves both methods on Panoptic.
Notably, our method achieves a substantially larger gain (MPJPE: $-20.8$) compared to HeatFormer (MPJPE: $-8.8$), a 2.4$\times$ improvement.
This is because our model is explicitly designed for multi-view world-coordinate reasoning with geometry-aware 3D fusion, which better absorbs the dense multi-view supervision and occlusion diversity provided by MVMP-HMR.
These results confirm that our synthetic benchmark provides complementary supervision for real-world multi-person HMR and that our model is the stronger beneficiary.

\section{Conclusion}

In this paper, we propose MVMP-HMR, a large multiview multi-person HMR benchmark and an end-to-end whole-body mesh recovery model for large occluded scenes.
As far as we know, this is the first study on multiview multi-person HMR and the first large MVMP-HMR benchmark in this area.
The proposed benchmark provides crowded multiview scenes with world-coordinate SMPL-X annotations, 3D joints, masks, and depth maps.
Based on this benchmark, our model constructs a scene-level 3D feature volume, extracts person-specific queries from predicted pelvis locations, and decodes SMPL-X parameters through a human transformer block.
Moreover, we introduce two novel losses, i.e., the orientation loss and the 3D joint density loss, to alleviate orientation and pose ambiguities under severe occlusions.
Experiments demonstrate that MVMP-HMR handles large-scale occluded multi-person scenes more effectively than existing HMR and multiview HPE methods.

\textbf{Limitations and future work.}
Our current setting requires calibrated multiview cameras with known intrinsic and extrinsic parameters.
Although accurate calibration can be challenging in real-world deployment, existing multiview matching and reconstruction methods such as MVS~\cite{schoenberger2016mvs} and VGGT~\cite{wang2025vggt} can estimate camera parameters from multiple views.
Future work will explore multiview multi-person HMR under weaker calibration assumptions and further improve domain transfer from synthetic data to real-world scenes.

\small{
\section*{Acknowledgements}
This work was supported by National Key R\&D Program of China (2024YFB3908500, 2024YFB3908504), NSFC (62202312), ICFCRT (W2441020), Shenzhen Science and Technology Program (KJZD20240903100022028, KQTD20210811090044003), Scientific Development Fund from Guangdong Provincial Key Laboratory of Visual Media and Multidimensional Intelligence, and Scientific Foundation for Youth Scholars of Shenzhen University.}

    % Human mesh recovery (HMR) refers to recovering the human 3D meshes in images. Most existing HMR tasks focus on multi-persons from a single image or a single person from multiple views. And the evaluation benchmarks used in these methods usually contain quite small numbers of humans or under small scenes, which is unreliable for real applications with severe occlusions. Thus, we present Multiview-HMR, a multiview model for multi-person whole-body human mesh recovery from multi-view images under occluded scenes. Specifically, Multiview-HMR first fuses multiple views to obtain a 3D feature volume for all persons, and then the pelvis joint from a 3D pose estimation net is utilized to acquire the human query of each person on the 3D feature volume. Finally, the human queries are cross-attentioned with the 3D feature volume and integrated to decode each person's 3D meshes. Besides, two novel losses are put forward to further enhance the model performance: the orientation loss and the 3D joint density loss, dealing with the orientation and pose ambiguities in the mesh predictions under the occluded scenes.
    % Furthermore, a large synthetic multiview HMR dataset is proposed, which consists of 15 multiview scenes with up to 50 camera views and 30 persons in each scene. Experiments demonstrate that the existing state-of-the-art (SOTA) HMR methods cannot perform well on the proposed large multiview HMR benchmark and the proposed Multiview-HMR model's advantages over existing SOTAs under large scenes with severe occlusions.

\bibliographystyle{IEEEtran}
\bibliography{egbib}

\vfill

\end{document}